\documentclass{article}
\usepackage[T1]{fontenc}
\usepackage{iclr2027_conference,times}

\usepackage{amsmath,amsfonts,bm}

\def\eqref#1{equation~\ref{#1}}

\def\1{\bm{1}}

\DeclareMathAlphabet{\mathsfit}{\encodingdefault}{\sfdefault}{m}{sl}
\SetMathAlphabet{\mathsfit}{bold}{\encodingdefault}{\sfdefault}{bx}{n}

\usepackage{url}
\usepackage{algorithm}
\usepackage{algpseudocode}
\usepackage{graphicx}
\usepackage{booktabs}
\usepackage{multirow}
\usepackage{tabularx}
\usepackage{xcolor}
\usepackage[most]{tcolorbox}
\usepackage{hyperref}

\newtcblisting{promptbox}[1]{
  enhanced,
  breakable,
  listing only,
  title={#1},
  colback=blue!2,
  colframe=blue!55!black,
  colbacktitle=blue!45!black,
  coltitle=white,
  fonttitle=\bfseries\small,
  boxrule=0.6pt,
  arc=2mm,
  boxsep=0.8mm,
  left=2.2mm,
  right=2.2mm,
  top=1.2mm,
  bottom=1.2mm,
  before skip=6pt,
  after skip=8pt,
  listing options={
    basicstyle=\rmfamily\fontsize{8.2}{9.8}\selectfont,
    breaklines=true,
    breakatwhitespace=true,
    breakautoindent=false,
    breakindent=0pt,
    columns=fullflexible,
    keepspaces=true,
    showstringspaces=false,
    upquote=true,
    tabsize=4,
    xleftmargin=0pt,
    xrightmargin=0pt,
    aboveskip=0pt,
    belowskip=0pt
  }
}

\newtcblisting{codebox}[1]{
  enhanced,
  breakable,
  listing only,
  title={#1},
  colback=green!2,
  colframe=green!45!black,
  colbacktitle=green!35!black,
  coltitle=white,
  fonttitle=\bfseries\small,
  boxrule=0.6pt,
  arc=2mm,
  boxsep=0.7mm,
  left=1.8mm,
  right=1.8mm,
  top=1mm,
  bottom=1mm,
  before skip=5pt,
  after skip=6pt,
  listing options={
    basicstyle=\ttfamily\fontsize{7}{8}\selectfont,
    breaklines=true,
    breakatwhitespace=true,
    breakautoindent=false,
    breakindent=0pt,
    columns=fullflexible,
    keepspaces=true,
    showstringspaces=false,
    upquote=true,
    tabsize=4,
    xleftmargin=0pt,
    xrightmargin=0pt,
    aboveskip=0pt,
    belowskip=0pt
  }
}

\title{SPO: Discovering Adaptive Large Neighborhood Search  Operators via Stackelberg Program Optimization}

\author{%
Xinyi Ke\textsuperscript{1,2} \quad
Kai Li\textsuperscript{1,2}\thanks{Corresponding author: \texttt{kai.li@ia.ac.cn}} \quad
Junliang Xing\textsuperscript{3} \quad
Yifan Zhang\textsuperscript{1,2,5} \quad
Jian Cheng\textsuperscript{1,2,4}
\\[0.6em]
\normalfont\small
\textsuperscript{1}C\textsuperscript{2}DL, Institute of Automation, Chinese Academy of Sciences
\\
\textsuperscript{2}School of Artificial Intelligence, University of Chinese Academy of Sciences
\\
\textsuperscript{3}Tsinghua University \quad
\textsuperscript{4}AiRiA
\\
\textsuperscript{5}University of Chinese Academy of Sciences, Nanjing
}

\iclrfinalcopy

\begin{document}

\maketitle
\fancyhead{}
\fancyhead[C]{\footnotesize SPO: Discovering Adaptive LNS Operators via Stackelberg Program Optimization}
\renewcommand{\headrulewidth}{0.4pt}

\begin{abstract}
Large neighborhood search (LNS) relies critically on destroy and repair operators, whose effectiveness depends on both adaptation to the evolving LNS state and interaction between the two roles. We introduce Stackelberg Program Optimization (SPO), an LLM-based framework for discovering adaptive executable destroy--repair programs. SPO conditions operator decisions on a compact LNS state, allowing state-dependent behavior to emerge through program discovery,
and organizes destroy--repair discovery as a Stackelberg interaction over program space that reflects their asymmetric dependency. Role-specific credits evaluate destroy programs as leaders and repair programs as conditional follower responses, guiding a coupled optimization process that combines LLM generator learning with population-based evolutionary search over programs. Experiments on the traveling salesperson problem and capacitated vehicle routing problem show that SPO outperforms strong baselines across a broad range of settings and generalizes beyond the discovery scale to larger instances and benchmark sets.
Behavioral analyses further demonstrate state-dependent operator behavior and coupled destroy--repair improvement during discovery.
\end{abstract}

\section{Introduction}

Combinatorial optimization problems arise widely in routing, scheduling, planning, and resource allocation, yet their rapidly growing search spaces often make exact optimization computationally prohibitive. Large neighborhood search (LNS) has emerged as a powerful metaheuristic framework for obtaining high-quality solutions within practical computational budgets. It repeatedly destroys part of a current solution and repairs the resulting partial structure, enabling broad exploration while preserving useful solution components \citep{shaw1998constraint,pisinger2018lns,mara2022survey}. Its effectiveness, however, depends critically on the design of its destroy and repair operators.

Effective LNS operator discovery should account for two intrinsic properties: \textbf{adaptation} over the LNS search process and \textbf{interaction} between destroy and repair, rather than treating operators as fixed and isolated rules. As LNS progresses, the strategy that is effective at one stage may no longer be appropriate at another.
For example, an operator may preserve the current solution more carefully while progress is being made, but change how it modifies the solution once the LNS search stagnates.
Meanwhile, destroy and repair are inherently coupled: the neighborhood induced by destruction shapes the reconstruction problem faced by repair, making the effectiveness of either dependent on the other. This raises a central question: \emph{how can we discover LNS operators that adapt to the evolving LNS state while coordinating destroy and repair?}

Designing such adaptation by hand is difficult: beyond specifying what an operator should do, one must determine how its strategy should respond to the evolving LNS state. Recent advances in LLM--based algorithm discovery offer a natural alternative by enabling complex algorithmic behaviors to be discovered directly as executable programs \citep{romeraparedes2024funsearch,liu2024eoh,ye2024reevo}. Building on this capability, we represent LNS operators as \emph{adaptive operator programs} whose decisions are conditioned on the evolving LNS state. In this way, adaptation becomes part of the operator program itself, allowing both the operator strategy and its state-dependent changes during an LNS rollout to emerge through discovery rather than be prescribed by hand.

Interaction fundamentally changes how destroy and repair programs should be discovered: their effectiveness is coupled, yet their roles are asymmetric. We therefore introduce \emph{Stackelberg Program Optimization} (SPO), which formulates their discovery as a leader--follower interaction \citep{stackelberg1934marktform}. Rather than optimizing the two roles independently, SPO carries this interaction throughout program generation, evaluation, and optimization. In each program-discovery round, destroy candidates are generated first, followed by conditional repair responses, and the resulting program pairs are evaluated through LNS rollouts, with their performance translated into role-specific credits for the leader and follower programs.
These credits drive a coupled generator--population optimization process: we use Group Relative Policy Optimization (GRPO) \citep{shao2024deepseekmath} to update the role-specific LLM generators, while evolutionary selection and variation update the program populations. Together, they shape program generation in subsequent rounds.

Experiments on the traveling salesperson problem (TSP) and capacitated vehicle routing problem (CVRP) show that SPO discovers high-quality destroy--repair programs and generalizes from the discovery setting to larger instances and benchmark distributions. Behavioral analyses provide evidence that the discovered operators exhibit adaptive behavior in response to the evolving LNS state, while destroy and repair develop in a coupled manner during discovery. Controlled ablations confirm the contributions of Stackelberg credit assignment and state conditioning. Together, these results show that SPO provides a framework for discovering adaptive and interacting LNS operators.

\section{Related Work}
\subsection{Large Neighborhood Search}

LNS explores large implicit neighborhoods through iterative destruction and reconstruction \citep{shaw1998constraint,pisinger2018lns,mara2022survey}. Classical methods use handcrafted destroy and repair heuristics, such as related removal and regret-based insertion \citep{shaw1998constraint,ropke2006alns,pisinger2018lns,voigt2025operators}.
Learning-based LNS methods have introduced parametric models into components of LNS, including reconstruction \citep{hottung2020neural,falkner2022construction}, removal \citep{chen2020dynamic,hottung2025nds}, and neighborhood selection in integer programming \citep{song2020general,sonnerat2021learning,wu2021lns,huang2023cllns}. Another line of work focuses on controlling which operators are applied during LNS execution. ALNS selects among predefined operators using performance-based weights \citep{ropke2006alns}, while subsequent methods use bandit or reinforcement-learning approaches for operator selection and LNS search control \citep{reijnen2024online,cai2025balans}. Recent work considers dependencies between destroy and repair in operator selection and coordination \citep{yu2026dual}.

\subsection{LLM-Based Operator Discovery}

LLM-based automatic heuristic design searches directly over executable programs, using task performance as feedback to iteratively improve generated heuristics. EoH evolves heuristic programs through LLM-based generation and evolutionary operators, while FunSearch combines an LLM with an evaluator to iteratively search for high-performing programs \citep{liu2024eoh,romeraparedes2024funsearch}. ReEvo further incorporates reflective feedback from heuristic evaluations to guide subsequent evolution \citep{ye2024reevo}. Subsequent work has extended this paradigm along several directions, including more efficient heuristic-program discovery \citep{wu2025hercules},
co-evolution of solver programs and instance generators \citep{ke2026gametheoretic},
and joint optimization of heuristic programs and the LLM generator \citep{huang2025calm}.

Recent work has applied this paradigm to LNS components. LLM-LNS discovers neighborhood-selection strategies for large-scale MILPs, determining which variables are released for reoptimization \citep{ye2025llmlns}. For vehicle routing, VRPAgent discovers problem-specific destroy and reinsertion-ordering operators while retaining a fixed greedy insertion procedure \citep{hottung2025vrpagent}. G-LNS discovers both destroy and repair programs for LNS \citep{zhao2026glns}, with SpecAHD further specializing repair heuristics to different local regions in large-scale routing \citep{lai2026specahd}, while full-component evolution applies program discovery across multiple components of ALNS \citep{yu2026fullcomponent}.
SPO differs in how destroy and repair are represented and optimized: their programs condition their behavior on the evolving LNS state, while their discovery is organized through an asymmetric leader--follower formulation.

\section{Preliminaries}

\subsection{Large Neighborhood Search}

Consider a combinatorial optimization instance $I$ in minimization form, with feasible solution set $\mathcal{X}(I)$ and objective $f_I(x)$. At iteration $t$, LNS maintains an incumbent solution $x_t\in\mathcal{X}(I)$. Let $\xi_t$ collect the stochastic choices made by the operators. A destroy operator $D$ removes a subset of the incumbent's components and returns a partial solution $\tilde{x}_t$ together with the removed set $V_t$; a repair operator $R$ then reconstructs a feasible candidate $x'_t$:
\begin{equation}
    (\tilde{x}_t,V_t)=D(x_t;\xi_t),
    \qquad
    x'_t=R(\tilde{x}_t,V_t;\xi_t).
    \label{eq:lns-update}
\end{equation}
An acceptance rule uses the candidate $x'_t$ to determine the next incumbent $x_{t+1}$. Writing $\xi=(\xi_0,\ldots,\xi_{T-1})$, repeated updates produce the solution trajectory
\begin{equation}
    \tau(D,R;I,\xi)=(x_0,x_1,\ldots,x_T).
    \label{eq:lns-trajectory}
\end{equation}

\subsection{Stackelberg Games}

Stackelberg games \citep{stackelberg1934marktform} model sequential interactions in which two decisions are interdependent: a leader acts first and shapes the decision faced by a follower, while the follower's anticipated response in turn affects the leader's optimal decision, as in supplier--retailer pricing \citep{liu2007pricing} or defender--attacker resource allocation \citep{korzhyk2010complexity}.
Formally, let $U(s_L,s_F)$ denote the utility shared by the leader and follower. The follower selects a best response to the leader's strategy, and the leader optimizes while anticipating this response:
\begin{equation}
\begin{aligned}
s_F^\star(s_L)
&\in \arg\max_{s_F} U(s_L,s_F),\
s_L^\star
&\in \arg\max_{s_L}
U\bigl(s_L,s_F^\star(s_L)\bigr).
\end{aligned}
\label{eq:stackelberg-background}
\end{equation}
This formulation captures the asymmetric dependence between the two decisions.
\section{Stackelberg Program Optimization}

\begin{figure}[t]
    \centering
    \IfFileExists{figures/lns.drawio.pdf}{%
        \includegraphics[width=\linewidth]{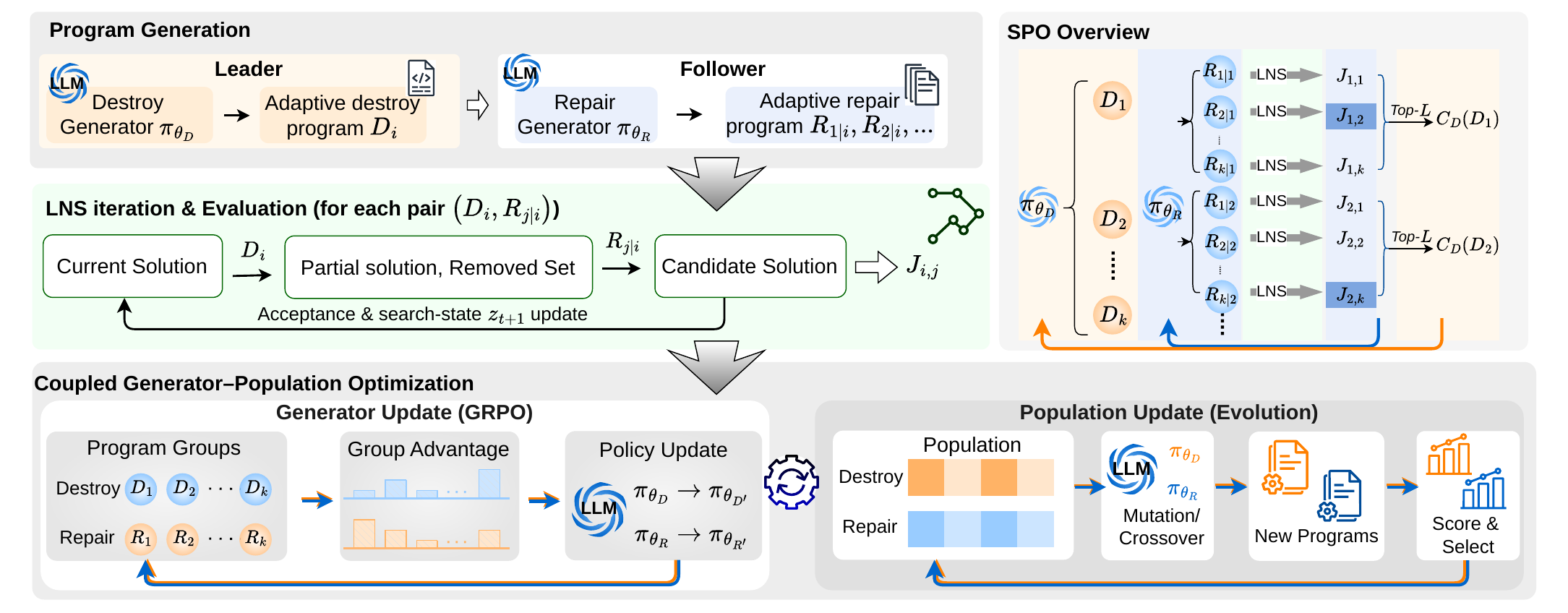}
    }{%
        \fbox{\parbox[c][35mm][c]{0.94\linewidth}{\centering
            Missing figure file\\[4pt]
            \texttt{\detokenize{figures/lns.drawio.pdf}}\\[4pt]
            Please restore the original figure.}}
    }
    \setlength{\abovecaptionskip}{1pt}
    \caption{Overview of Stackelberg Program Optimization (SPO). \textbf{Program Generation} produces leader destroy programs and conditionally generated follower repair programs. \textbf{LNS Iteration \& Evaluation} executes each program pair with LNS-state conditioning and evaluates its rollout utility. \textbf{Coupled Generator--Population Optimization} uses role-specific credits to update the LLM generators via GRPO and evolve the executable-program populations. \textbf{SPO Overview} summarizes the hierarchical generation, evaluation, and optimization process.}
    \label{fig:spo-overview}
\end{figure}

We formulate adaptive operator discovery as \emph{Stackelberg Program Optimization} (SPO). SPO maintains role-specific LLM generators $\pi_{\theta_D}$ and $\pi_{\theta_R}$, which generate executable destroy and repair programs whose decisions can depend on the evolving LNS state.
Destroy programs act as leaders and repair programs as followers; their joint evaluations yield role-specific credits that drive coupled generator--population optimization.
Figure~\ref{fig:spo-overview} summarizes the framework.

During program discovery, candidate destroy--repair program pairs are evaluated through LNS rollouts. The discovery process searches over executable programs, whereas each rollout performs the underlying solution search for a given instance. Throughout, \emph{program} refers to sampled executable code, while \emph{operator} refers to the destroy or repair function instantiated by a program within LNS. Within each rollout, the programs remain fixed, while their decisions may vary with the current solution and the LNS state. At test time, the selected programs are directly executed within LNS.

\subsection{Adaptive Operator Programs}

LNS is inherently history dependent: the same incumbent solution may call for different decisions depending on whether it has just been improved or has remained unchanged over many iterations. Let \(h_t\) denote the LNS history up to iteration \(t\), including previous solutions and improvement outcomes. We introduce a compact \emph{LNS state}
\begin{equation}
z_t=\phi(h_t),
\end{equation}
and provide this state as an input to both operator programs:
\begin{equation}
(\tilde{x}_t,V_t)=D(x_t,z_t;\xi_t),
\qquad
x'_t=R(\tilde{x}_t,V_t,z_t;\xi_t).
\label{eq:adaptive-operators}
\end{equation}

This formulation makes adaptation part of the operator discovery space. Rather than encoding a fixed heuristic, an adaptive program can jointly determine an operator strategy and how its decisions change with the LNS state. Equivalently, a program \(O(x,z)\) compactly represents a state-indexed family of operators \(\{O_z:x\mapsto O(x,z)\}_{z\in\mathcal Z}\),
whose effective strategy may vary with the LNS state.

The formulation is agnostic to the particular state representation \(\phi\). Ideally, the LNS state should provide trajectory information complementary to the current solution while remaining compact and avoiding unnecessary dependence on problem- or instance-specific quantities. We therefore deliberately use a minimal, problem-independent signal,
\begin{equation}
z_t=\text{number of iterations since the last improvement}.
\end{equation}
It is reset to zero after an improvement and incremented otherwise, providing a simple measure of recent progress versus increasing stagnation. This choice introduces negligible overhead and leaves the state-dependent decision logic itself to be discovered by the program. Richer representations could incorporate additional trajectory information, while their design and transferability across problems and instance distributions remain open directions.

\subsection{Stackelberg Formulation for Operator Discovery}

We formulate destroy--repair discovery as a Stackelberg interaction, with destroy programs acting as leaders and repair programs as followers. Reflecting this ordering, SPO first samples destroy candidates and then generates repair candidates conditionally on each destroy:
\begin{equation}
D_i\sim\pi_{\theta_D}(\cdot\mid\gamma_D),
\qquad
R_{j\mid i}\sim\pi_{\theta_R}(\cdot\mid\gamma_R,D_i),
\label{eq:conditional-program-generation}
\end{equation}
where \(\gamma_D\) and \(\gamma_R\) denote the corresponding generation contexts, specifying the operator role, program interface, generation constraints, and optional program context. Here \(R_{j\mid i}\) denotes the \(j\)-th repair candidate generated conditionally on destroy program \(D_i\).

To evaluate a destroy--repair pair, let \(\tau(D,R;I,\xi)\) denote its LNS rollout on instance \(I\), where \(\xi\) captures rollout stochasticity, and let
\begin{equation}
x_\tau^\star\in\arg\min_{x\in\tau(D,R;I,\xi)} f_I(x)
\end{equation}
be the best feasible solution encountered. We define the pair utility as
\begin{equation}
J(D,R)=\mathbb{E}_{I,\xi}
\left[
\frac{f_I(x_0)-f_I(x_\tau^\star)}
{\max{|f_I(x_0)|,\epsilon_{\mathrm{den}}}}
\right],
\qquad \epsilon_{\mathrm{den}}>0,
\label{eq:pair-performance}
\end{equation}
where the expectation is over training instances and stochastic rollouts, with higher values indicating greater improvement. The small constant \(\epsilon_{\mathrm{den}}\) stabilizes the denominator.

The follower best response to a destroy program is
\begin{equation}
R^\star(D)\in\arg\max_{R\in\mathcal{R}}J(D,R),
\label{eq:follower-response}
\end{equation}
and the leader anticipates this response:
\begin{equation}
D^\star\in\arg\max_{D\in\mathcal{D}}
J\bigl(D,R^\star(D)\bigr).
\label{eq:spo-objective}
\end{equation}
Since exact best responses cannot be computed over the space of executable programs, SPO realizes these objectives using sampled candidates and role-specific credit assignment.

\subsection{Stackelberg Credit Assignment}
\label{sec:credit}

The Stackelberg formulation induces distinct optimization signals for the two operator roles. A repair candidate is evaluated by how well it responds to the destroy program on which it is conditioned, whereas a destroy candidate is evaluated through the strong repair responses it elicits.

\textbf{Follower credit.}
For a destroy candidate \(D_i\), let
\(\mathcal{R}_i=\{R_{1\mid i},\ldots,R_{N\mid i}\}\)
denote its conditionally generated repair candidates. Following Equation~\ref{eq:follower-response}, each repair candidate receives the credit
\begin{equation}
C_R(R_{j\mid i};D_i)=J(D_i,R_{j\mid i}).
\label{eq:follower-credit}
\end{equation}

\textbf{Leader credit.}
Since the exact best response in Equation~\ref{eq:follower-response} is unavailable, we approximate it using the strongest conditionally sampled repair candidates. Let \(\operatorname{Top}_{L}(\mathcal{R}_i)\) contain the \(L\) candidates with the highest pair utilities. The leader credit is
\begin{equation}
C_D(D_i)=\frac{1}{L}
\sum_{R_{j\mid i}\in\operatorname{Top}_{L}(\mathcal{R}_i)}
J(D_i,R_{j\mid i}).
\label{eq:leader-credit}
\end{equation}
Averaging the top-\(L\) responses rather than taking a single empirical maximum reduces sensitivity to rollout noise while retaining emphasis on strong follower responses.

\subsection{Coupled Generator--Population Optimization}
\label{sec:optimization}
SPO jointly improves the two components governing program generation in Equation~\ref{eq:conditional-program-generation}: the role-specific generators \(\pi_{\theta_D}\) and \(\pi_{\theta_R}\), and the program populations that provide context through \(\gamma_D\) and \(\gamma_R\). GRPO optimizes the generators from role-specific credits, while evolutionary optimization updates the populations used as program context. These two processes are coupled across discovery rounds, as each generation is produced by the current generators under the current population contexts and subsequently used to update both.
Algorithm~\ref{alg} (Appendix~\ref{app:training-algorithm}) summarizes SPO training.

\textbf{GRPO-based generator update.}
We use Group Relative Policy Optimization (GRPO) \citep{shao2024deepseekmath} to optimize the role-specific generators based on the credits defined above. Specifically, \(C_D(D_i)\) is used for destroy programs and \(C_R(R_{j\mid i};D_i)\) for repair programs.
For either role, let \(\{y_i\}_{i=1}^{B}\) denote a group of programs sampled from \(\pi_{\theta_{\mathrm{old}}}(\cdot\mid\gamma)\) under the same generation context \(\gamma\).
Given their corresponding credits \(\{C_i\}_{i=1}^{B}\), we compute the group-centered advantage
\begin{equation}
    A_i=C_i-\frac{1}{B}\sum_{j=1}^{B}C_j.
    \label{eq:grpo-advantage}
\end{equation}
The probability ratio between the updated and sampling policies is
\begin{equation}
    \rho_i(\theta)
    =\frac{\pi_\theta(y_i\mid\gamma)}
    {\pi_{\theta_{\mathrm{old}}}(y_i\mid\gamma)}.
    \label{eq:grpo-ratio}
\end{equation}
The corresponding generator is updated by maximizing the clipped objective
\begin{equation}
    \mathcal{L}_{\mathrm{GRPO}}(\theta)
    =\mathbb{E}\!\left[
        \frac{1}{B}\sum_{i=1}^{B}
        \min\!\left(
            \rho_i(\theta)A_i,
            \operatorname{clip}\!\left(
                \rho_i(\theta),1-\epsilon,1+\epsilon
            \right)A_i
        \right)
    \right],
    \label{eq:grpo}
\end{equation}
where \(\epsilon>0\) controls clipping. This update is applied separately to the destroy and repair generators, \(\pi_{\theta_D}\) and \(\pi_{\theta_R}\). For repair updates, the sampling and probability-ratio expressions are conditioned on the same destroy program \(D_i\) within each group; this dependence is omitted for brevity.

\textbf{Evolutionary population update.}
We use evolutionary optimization to update the role-specific program populations that provide context for subsequent generation. At discovery round \(k\), let \(\mathcal{P}_D^{(k)}\) and \(\mathcal{P}_R^{(k)}\) denote the current destroy and repair populations. To generate new candidates, programs from these populations are incorporated into \(\gamma_D\) and \(\gamma_R\) as program context. The corresponding role-specific generators then generate mutations of individual programs or crossovers that combine multiple programs, producing candidate sets
\(\mathcal{G}_D^{(k)}\) and \(\mathcal{G}_R^{(k)}\).

Destroy programs can be compared using \(C_D\). Repair candidates, however,
may originate from different conditioning destroy programs, making their
conditional credits dependent on different leader contexts. To obtain a common
basis for repair population selection, we therefore construct a diverse leader
panel \(\mathcal{E}_D=\{D_1,\ldots,D_M\}\) and define the cross-play score
\begin{equation}
    C_R^{\mathrm{pop}}(R)
    =
    \frac{1}{M}\sum_{m=1}^{M}J(D_m,R).
    \label{eq:repair-population-credit}
\end{equation}

The populations are selected from
\(\mathcal{P}_D^{(k)}\cup\mathcal{G}_D^{(k)}\)
and
\(\mathcal{P}_R^{(k)}\cup\mathcal{G}_R^{(k)}\),
using \(C_D\) and \(C_R^{\mathrm{pop}}\), respectively.

\section{Experiments}

We evaluate SPO on the traveling salesperson problem (TSP) and capacitated vehicle routing problem (CVRP) to answer three questions: (i) whether SPO discovers adaptive operators that improve solution quality; (ii) whether the discovered operators generalize across instance sizes and distributions; and (iii) how Stackelberg credit assignment and LNS-state conditioning contribute to performance.

\subsection{Experimental Setup}

\textbf{Common settings. }
We use \texttt{Qwen3.5-9B} \citep{qwen3.5} as a shared backbone with role-specific LoRA adapters \citep{hu2021lora} for the destroy and repair generators.
SPO runs for 30 program-discovery rounds and maintains at most 10 executable programs per role. Discovery evaluates candidate pairs using 100-step LNS rollouts with non-worsening acceptance and retains the pair with the highest discovery utility \(J\) found so far. Final evaluation runs the retained pair for 500 LNS iterations.
See Appendices~\ref{app:implementation}, \ref{app:problems},
and~\ref{app:evaluation-protocol} for implementation and evaluation details.

\textbf{Evaluation metric. }
Discovery optimizes the relative-improvement utility $J$ in Equation~\ref{eq:pair-performance}. Final performance is measured by the reference gap $\operatorname{Gap}(x)=100(f(x)-f^{\mathrm{ref}})/f^{\mathrm{ref}}$, using benchmark optimal or best-known values when available and high-quality solver solutions otherwise.
Results are averaged over test instances; lower is better.

\subsubsection{Traveling Salesperson Problem}

\textbf{Setting and benchmarks. }
SPO is trained on uniformly sampled 50-node Euclidean TSP instances. We evaluate the discovered operators on held-out uniform instances of increasing size and on TSPLIB \citep{reinelt1991tsplib}, which provides benchmark instances with more heterogeneous geometric structure than the synthetic training distribution. Reference values for synthetic instances are obtained with Concorde \citep{applegate2011tsp}, while TSPLIB uses the optimal tour lengths.

\textbf{Baselines. }
We consider two baseline groups. Handcrafted methods include the construction heuristics Nearest Neighbor (NN) and Farthest Insertion (FI), the local-search methods 2-opt and 3-opt \citep{rosenkrantz1977analysis,croes1958method,lin1965computer}, and ALNS \citep{ropke2006alns}, which adaptively selects from a fixed operator portfolio. Program-discovery methods include the general LLM-based frameworks EoH \citep{liu2024eoh}, FunSearch \citep{romeraparedes2024funsearch}, and ReEvo \citep{ye2024reevo}, together with G-LNS \citep{zhao2026glns}, which specifically discovers destroy and repair programs for LNS.
All program-discovery baselines use the same final LNS iteration budget as SPO; EoH, FunSearch, and ReEvo also use SPO's destroy--repair interface
(Appendix~\ref{app:evaluation-protocol}).

\subsubsection{Capacitated Vehicle Routing Problem}

\textbf{Setting and benchmarks. }
SPO is trained on synthetic 50-customer CVRP instances with uniformly sampled coordinates and demands. We evaluate the discovered operators on held-out uniform instances and on the A, B, P, and X families from CVRPLIB \citep{augerat1995computational,uchoa2017benchmark}, which cover benchmark instances with varied spatial, demand, and capacity characteristics.

\textbf{Baselines. }
The handcrafted and solver baselines comprise Clarke--Wright savings (CW) \citep{clarke1964scheduling}, Google OR-Tools \citep{ortools2024}, local search (LS) using relocate, swap, and 2-opt neighborhoods \citep{vidal2014unified}, and ALNS \citep{ropke2006alns}. We use the same set of program-discovery baselines where applicable: EoH, FunSearch, ReEvo, and G-LNS.

\begin{table}[t]
\caption{Mean final reference gaps (\%) on TSP and CVRP benchmarks. Bold
indicates the best reported result in each column; lower is better.}
\label{tab:main_results}
\centering
\footnotesize
\setlength{\tabcolsep}{2.5pt}
\renewcommand{\arraystretch}{1.08}
\begin{tabularx}{\linewidth}{@{}l*{7}{>{\raggedleft\arraybackslash}X}@{}}
\toprule
\multicolumn{8}{@{}l}{\textbf{(a) Traveling Salesperson Problem}} \\
\cmidrule(lr){1-8}
\multirow{2}{*}{Method}
& \multicolumn{4}{c}{Uniform instances (number of nodes $n$)}
& \multicolumn{3}{c}{TSPLIB (instance-size range)} \\
\cmidrule(lr){2-5}\cmidrule(lr){6-8}
& $50$ & $100$ & $200$ & $500$
& $50$--$100$ & $100$--$200$ & $>200$ \\
\midrule
NN
& 21.295 & 23.807 & 24.884 & 25.999
& 26.039 & 23.143 & 25.710 \\
FI
& 6.261 & 7.711 & 8.909 & 10.535
& 7.336 & 7.490 & 11.662 \\
\addlinespace[1.5pt]
2-opt
& 3.796 & 4.197 & 5.078 & 5.179
& 4.120 & 3.707 & 5.600 \\
3-opt
& 2.232 & 2.905 & 3.900 & 4.578
& 2.859 & 3.041 & 4.764 \\
ALNS
& 1.720 & 2.370 & 3.106 & 4.103
& 1.865 & 2.414 & 4.106 \\
\specialrule{0.3pt}{1.5pt}{1.5pt}
EoH
& 1.222 & 4.582 & 8.843 & 15.708
& 2.305 & 3.859 & 11.627 \\
FunSearch
& 1.201 & 2.372 & 5.717 & 18.016
& 1.670 & 2.045 & 6.445 \\
ReEvo
& 1.252 & 3.707 & 6.642 & 11.577 & 2.210 & 3.869 & 11.166 \\
G-LNS
& 0.483 & 1.385 & 2.589 & 6.305
& 0.597 & 1.196 & 4.145 \\
\specialrule{0.3pt}{1.5pt}{1.5pt}
\textbf{SPO}
& \textbf{0.212} & \textbf{0.642} & \textbf{1.274} & \textbf{3.139}
& \textbf{0.408} & \textbf{0.574} & \textbf{4.040} \\
\bottomrule
\end{tabularx}

\vspace{4pt}

\setlength{\tabcolsep}{2.5pt}
\begin{tabularx}{\linewidth}{@{}l*{8}{>{\raggedleft\arraybackslash}X}@{}}
\toprule
\multicolumn{9}{@{}l}{\textbf{(b) Capacitated Vehicle Routing Problem}} \\
\cmidrule(lr){1-9}
\multirow{2}{*}{Method}
& \multicolumn{4}{c}{Uniform instances (number of customers $n$)}
& \multicolumn{4}{c}{CVRPLIB (instance family)} \\
\cmidrule(lr){2-5}\cmidrule(lr){6-9}
& $50$ & $100$ & $150$ & $200$
& A & B & P & X \\
\midrule
CW
& 5.929 & 6.418 & 6.271 & 6.056
& 5.089 & 4.416 & 8.102 & 6.013 \\
OR-Tools
& 6.036 & 7.245 & 9.811 & 10.902
& 4.626 & 4.183 & 4.401 & 7.755 \\
\addlinespace[1.5pt]
LS
& 5.450 & 5.590 & 5.752 & 5.218
& 5.121 & 5.793 & 7.706 & 5.529 \\
ALNS
& 5.892 & 6.000 & 6.956 & 7.165
& 3.979 & 3.486 & 3.910 & 9.680 \\
\specialrule{0.3pt}{1.5pt}{1.5pt}
EoH
& 3.842 & 7.497 & 12.252 & 15.788
& 4.451 & 3.811 & 3.607 & 16.199 \\
FunSearch
& 3.855 & 6.092 & 13.142 & 16.251
& 3.992 & 4.444 & 2.853 & 16.279 \\
ReEvo
& 5.737 & 8.531 & 10.616 & 14.870
& 5.088 & 5.388 & 3.587 & 15.747 \\
G-LNS
& 7.704 & 8.206 & 7.357 & 7.535
& 3.634 & 2.542 & 4.030 & 5.266 \\
\specialrule{0.3pt}{1.5pt}{1.5pt}
\textbf{SPO}
& \textbf{3.190} & \textbf{3.221} & \textbf{3.273}
& \textbf{4.093} & \textbf{2.445} & \textbf{1.918}
& \textbf{2.351} & \textbf{5.008} \\
\bottomrule
\end{tabularx}
\end{table}

\subsection{Main Results}

Table~\ref{tab:main_results} reports mean final reference gaps on TSP and CVRP. All LNS-based methods are evaluated with the same budget of 500 destroy--repair iterations.

\textbf{TSP. }
SPO achieves the lowest reference gap in all seven TSP groups. Its advantage generally widens beyond the $n=50$ discovery setting as several program-discovery baselines degrade more strongly with instance size, and remains strongest up to $n=500$. It also leads across all three TSPLIB ranges, showing that the discovered operators transfer beyond the synthetic discovery distribution.

\textbf{CVRP. }
CVRP provides a complementary test under capacity constraints and multi-route solution structure. SPO achieves the lowest gap in every group despite substantial variation in the strongest competing method across synthetic sizes and CVRPLIB families. On synthetic instances, its performance degrades much more slowly with scale than several program-discovery baselines, and the advantage also transfers across all four CVRPLIB families.

\begin{figure}[t]
    \centering
    \IfFileExists{figures/optimization_dynamics_crossplay.pdf}{%
        \includegraphics[width=\linewidth]{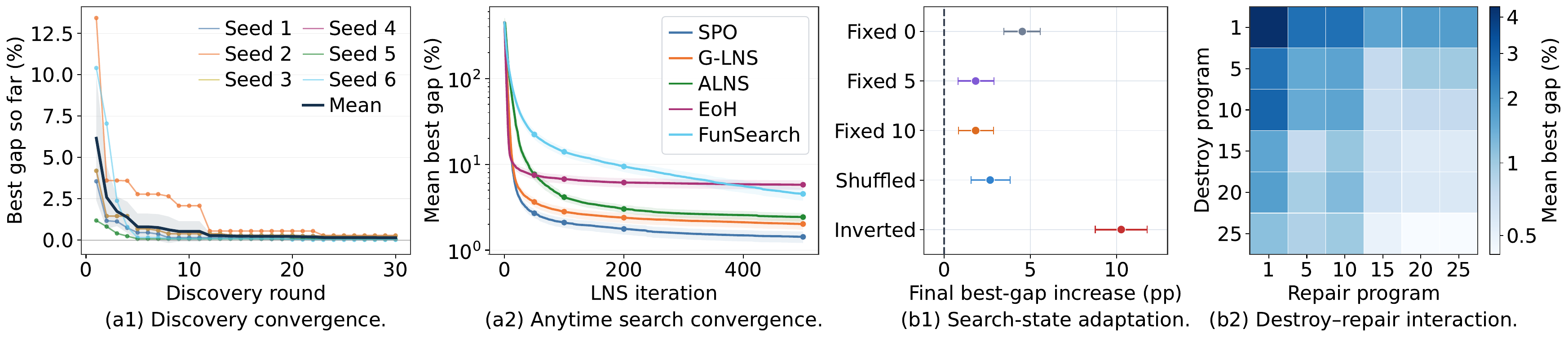}
    }{%
        \fbox{\parbox[c][35mm][c]{0.94\linewidth}{\centering
            Missing figure file\\[4pt]
            \texttt{\detokenize{figures/optimization_dynamics_crossplay.pdf}}\\[4pt]
            Please restore the original figure.}}
    }
    \setlength{\abovecaptionskip}{1pt}
    \caption{
    \textbf{Optimization dynamics and behavioral analysis of SPO.}
    (a1) Best-so-far discovery gap across six TSP runs; thick line: mean, shading: across-seed variation.
    (a2) Anytime best-gap convergence during TSP LNS execution.
    (b1) State interventions on one discovered TSP pair: \emph{Fixed $k$} sets $z_t=k$; \emph{Shuffled} permutes the observed state sequence; \emph{Inverted} maps $z_t<10$ to $30$ and otherwise to $0$. Shown are mean final-gap increases over unmodified execution with 95\% bootstrap CIs from 2{,}000 instance--seed resamples.
    (b2) Pairwise combinations of destroy and repair programs from the indicated rounds of one CVRP discovery run; colors show mean best gap (power norm \(\gamma=0.5\)).
    Lower is better; pp = percentage points.
    }
    \label{fig:optimization-dynamics}
\end{figure}

\subsection{Optimization Dynamics}

Figure~\ref{fig:optimization-dynamics}(a1) tracks the best-so-far discovery gap across six independent runs with different random seeds. The runs consistently improve over discovery rounds and converge to similarly low gaps, indicating stable discovery performance across seeds.
Panel (a2) compares LNS convergence using SPO's discovered operators with that of the baselines. SPO attains lower gaps early and continues improving after several baselines have largely plateaued, showing that its advantage persists throughout LNS execution rather than emerging only at the final evaluation point.

\begin{figure}[t]
    \centering
    \includegraphics[width=\linewidth]{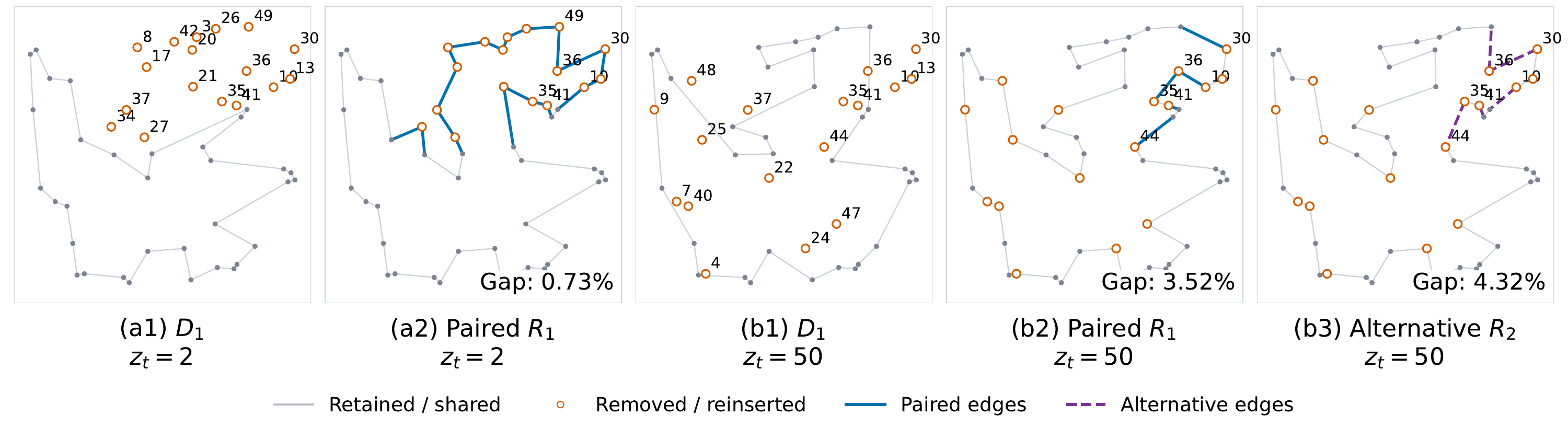}
    \setlength{\abovecaptionskip}{1pt}
    \caption{
    \textbf{Illustrative case study of state adaptation and destroy--repair interaction.}
    The same TSP instance and incumbent are used throughout.
    (a1--a2) Destroy and subsequent repair by a discovered pair $(D_1,R_1)$
    under the true state $z_t=2$.
    (b1--b2) The same pair under the counterfactual state $z_t=50$.
    (b3) Starting from the same destruction as in (b1), the paired repair
    $R_1$ is replaced by an alternative repair $R_2$, yielding a different
    reconstruction.
    }
    \label{fig:qualitative-behavior}
\end{figure}

\subsection{Behavioral Analysis}

\textbf{LNS-state adaptation. }
Figure~\ref{fig:optimization-dynamics}(b1) shows that disrupting the correspondence between the LNS state and search progress consistently degrades performance, with inversion producing the largest effect. This indicates that the tested pair makes meaningful use of the state signal rather than treating it as an incidental input.
Figure~\ref{fig:qualitative-behavior} provides a qualitative TSP case study of state adaptation. At a trajectory point with true state $z_t=2$, a discovered pair produces the destroy and repair decisions in (a1--a2). Keeping the instance and incumbent fixed, we counterfactually set $z_t=50$ and rerun the same pair in (b1--b2). The destroy pattern shifts from a relatively localized region of the tour to a more dispersed one, leading to a different reconstruction and illustrating a state-dependent transition from local refinement to broader exploration.

\textbf{Destroy--repair interaction. }
Figure~\ref{fig:optimization-dynamics}(b2) shows coupled improvement across discovery. While performance improves along both role axes, programs developed at similar stages also tend to form locally strong combinations, suggesting that destroy and repair are progressively optimized in relation to one another.
Figure~\ref{fig:qualitative-behavior}(b1--b3) provides an operator-level view of this interaction. The destruction in (b1) fixes the partial tour and removed nodes, thereby defining the subproblem faced by repair. Keeping this destroy output unchanged, replacing the paired $R_1$ with $R_2$ leads to a different reconstruction and a higher gap in this instance. This illustrates the conditional nature of destroy--repair interaction: the effect of a repair program depends on the particular partial structure produced by destroy, rather than being independent of its partner.

\subsection{Ablation Study}
\textbf{Stackelberg credit assignment.} We isolate the contribution of Stackelberg credit by replacing the role-specific leader--follower credits with a shared pair-level credit, where both programs in a destroy--repair pair receive the same rollout score.
The \textbf{w/o Stackelberg Credit} variant consistently degrades across uniform and TSPLIB instances (Table~\ref{tab:tsp-ablation}), showing that coupling the two programs through joint evaluation alone is not sufficient. Explicitly distinguishing their leader and follower roles provides a more effective credit signal for optimizing interacting destroy and repair programs.

\textbf{LNS-state adaptation.} We then remove the LNS-state input while preserving the Stackelberg credit structure. As shown in Table~\ref{tab:tsp-ablation}, the resulting \textbf{w/o State Conditioning} variant remains close to SPO around the discovery scale, but the performance gap widens as the evaluation scale grows. The same pattern extends to larger TSPLIB instances. This suggests that state conditioning is less important near the discovery setting itself, but becomes increasingly important for maintaining effective operator behavior as the discovered programs are transferred beyond it.

\begin{table}[t]
\caption{Controlled TSP ablations of Stackelberg credit assignment and
LNS-state adaptation. Values are mean final reference gaps (\%); lower is better.}
\label{tab:tsp-ablation}
\centering
\footnotesize
\setlength{\tabcolsep}{2.5pt}
\renewcommand{\arraystretch}{1.08}

\begin{tabularx}{\linewidth}{@{}l*{7}{>{\raggedleft\arraybackslash}X}@{}}
\toprule
\multirow{2}{*}{Variant}
& \multicolumn{4}{c}{Uniform instances (number of nodes $n$)}
& \multicolumn{3}{c}{TSPLIB (instance-size range)} \\
\cmidrule(lr){2-5}\cmidrule(lr){6-8}
& $50$ & $100$ & $200$ & $500$
& $50$--$100$ & $100$--$200$ & $>200$ \\
\midrule

SPO
& \textbf{0.212} & \textbf{0.642} & \textbf{1.274}
& \textbf{3.139} & \textbf{0.408} & \textbf{0.574} & \textbf{4.040} \\

w/o Stackelberg Credit
& 1.398 & 1.908 & 3.089
& 5.995 & 1.309 & 2.054 & 6.647 \\

w/o State Conditioning
& 0.273 & 0.816 & 2.686
& 6.484 & 0.592 & 0.962 & 6.850 \\

\bottomrule
\end{tabularx}
\end{table}
\section{Conclusion}

We introduced Stackelberg Program Optimization (SPO) for discovering adaptive operators in large neighborhood search. SPO represents destroy and repair operators as executable programs whose behavior can adapt over the course of LNS execution, and organizes their discovery through a leader--follower formulation that captures the asymmetric dependency between the two roles. Role-specific credit assignment, together with coupled generator--population optimization, guides the discovery of interacting operator programs. Experiments on TSP and CVRP show that the discovered operators achieve strong solution quality across instance sizes and benchmark distributions, while the ablation and behavioral analyses provide evidence for adaptive operator behavior and coupled destroy--repair improvement during discovery. Together, these results demonstrate the value of accounting for both search dynamics and inter-program dependencies in automated operator discovery.

Although we instantiate SPO for destroy--repair discovery, the formulation provides a general optimization framework for other systems with leader--follower dependencies. Extending SPO beyond two roles to multi-component or multi-stage systems is a natural direction toward more general structured program optimization. More broadly, our adaptive formulation suggests expanding program discovery from optimizing isolated algorithmic components toward optimizing more of the algorithmic system itself. Mechanisms that are traditionally specified outside an individual program---such as when and how its strategy should change during execution---can instead be incorporated into the program search space and discovered jointly with the underlying algorithmic logic. This opens a direction toward system-level algorithm discovery, where both component behavior and the mechanisms governing its evolution during execution are optimized within a unified search space.

\bibliography{main}
\bibliographystyle{iclr2027_conference}

\appendix
\section{Program Discovery and Training Details}
\label{app:implementation}
This section provides the training procedure and implementation details of SPO; problem-specific settings and evaluation protocols are described in Appendices~\ref{app:problems} and~\ref{app:evaluation-protocol}, respectively.

\subsection{Overall Training Algorithm}
\label{app:training-algorithm}

Algorithm~\ref{alg} summarizes the complete SPO training procedure. In each
discovery round, SPO conditionally generates and evaluates destroy--repair
programs, updates the role-specific generators with GRPO, and evolves the two
program populations.
Throughout discovery, SPO retains the evaluated destroy--repair pair with the highest utility \(J\) found so far and returns it after \(K\) rounds.

\begin{algorithm}[htbp]
\caption{Stackelberg Program Optimization}
\label{alg}
\begin{algorithmic}[1]
\Require Role-specific LLM generators
\(\pi_{\theta_D}, \pi_{\theta_R}\)
\State Initialize program populations
\(\mathcal{P}_D^{(1)}, \mathcal{P}_R^{(1)}\)

\For{program-discovery round \(k=1,\ldots,K\)}
    \State Construct destroy generation contexts from
    \(\mathcal{P}_D^{(k)}\)
    \State Generate destroy candidates
    \(\mathcal{G}_D^{(k)}\) using \(\pi_{\theta_D}\)

    \ForAll{\(D_i \in \mathcal{G}_D^{(k)}\)}
        \State Construct repair generation contexts from
        \(\mathcal{P}_R^{(k)}\), conditioned on \(D_i\)
        \State Generate repair candidates
        \(\mathcal{R}_i\) using \(\pi_{\theta_R}\)
        \State Evaluate \(J(D_i,R_{j\mid i})\) for all
        \(R_{j\mid i}\in\mathcal{R}_i\) and compute \(C_D(D_i)\)
        using Equation~\ref{eq:leader-credit}
    \EndFor

    \State \(\mathcal{G}_R^{(k)}
    \gets \bigcup_{D_i\in\mathcal{G}_D^{(k)}} \mathcal{R}_i\)

    \State Update the best-so-far pair using the pairs evaluated in round \(k\)

    \State Center \(C_D(D_i)\) and \(J(D_i,R_{j\mid i})\) within their
    generation-context groups
    \State Update \(\pi_{\theta_D}\) and \(\pi_{\theta_R}\) with GRPO

    \State Construct a diverse leader panel
    \(\mathcal{E}_D^{(k)}\)
    \State Evaluate cross-play score
    \(C_R^{\mathrm{pop}}(R)\) for repair population candidates

    \State Select \(\mathcal{P}_D^{(k+1)}\) from
    \(\mathcal{P}_D^{(k)}\cup\mathcal{G}_D^{(k)}\)
    using \(C_D\) with diversity-aware survivor selection
    \State Select \(\mathcal{P}_R^{(k+1)}\) from
    \(\mathcal{P}_R^{(k)}\cup\mathcal{G}_R^{(k)}\)
    using \(C_R^{\mathrm{pop}}\) with diversity-aware survivor selection
\EndFor

\State Select the retained best-so-far pair \((D_{\mathrm{sel}},R_{\mathrm{sel}})\)
\State \Return selected pair \((D_{\mathrm{sel}},R_{\mathrm{sel}})\) and
populations \(\mathcal{P}_D^{(K+1)},\mathcal{P}_R^{(K+1)}\)
\end{algorithmic}
\end{algorithm}

\subsection{Model, Adapters, and Sampling}
\label{app:model-configuration}

SPO uses \texttt{Qwen3.5-9B} as a frozen shared backbone with separate
trainable LoRA adapters for the destroy and repair roles. Program generation
uses vLLM with the corresponding role-specific adapter, while adapter updates
are performed separately in PyTorch. Generation and training run sequentially
on a single NVIDIA A100-SXM4 GPU with 40\,GB of memory.

The maximum context length is 8,192 tokens. Programs are sampled with
temperature \(0.8\) and top-\(p\) \(0.95\), with maximum generation lengths of
1,200 tokens for TSP and 2,048 tokens for CVRP. All adapters use rank 16,
scaling factor 32, zero dropout, and bfloat16 computation. The TSP adapters
target the query, key, value, and output projections, while the CVRP adapters
additionally target the gate, up, and down projections.

\subsection{Discovery Budget and Runtime}
\label{app:discovery-budget}

Each discovery round generates 30 destroy candidates from three mutation and
two crossover groups, with six responses per group. For each valid destroy,
SPO conditionally generates up to 30 repair candidates, yielding up to 900
destroy--repair pairs per round. Over 30 rounds, the nominal upper budget is
900 destroy and 27,000 repair candidates. Candidate pairs are evaluated using
100-step LNS rollouts; validation and final evaluation use 500-step rollouts.

A representative TSP discovery run generated 900 destroy and 17,850 repair
programs and evaluated approximately 14,500 program pairs. It took
approximately 17.3 hours, with rollout evaluation parallelized across 160
workers on Intel Xeon Platinum 8358 CPUs.

\subsection{Role-Specific Credits and Generator Updates}
\label{app:grpo-details}

The destroy and repair adapters use independent AdamW optimizers
\citep{loshchilov2017decoupled}, each with learning rate
\(5\times10^{-6}\), clipping radius \(\epsilon=0.2\), and maximum gradient
norm \(1.0\). The optimization micro-batch sizes are 3 for TSP and 2 for CVRP;
gradients are accumulated over all usable samples for each role before clipping
and a single optimizer step.

The follower best response is approximated using conditionally generated repair
programs. For each destroy program \(D_i\), the leader credit \(C_D(D_i)\)
averages the utilities of its top \(L=2\) sampled repair responses, while each
repair program receives the conditional follower credit
\(C_R(R_{j\mid i};D_i)=J(D_i,R_{j\mid i})\).

For both roles, advantages are centered within each group of programs sampled
from the same generation context. Samples without a finite credit or usable
policy log-probability are omitted from the update. The resulting advantages
are optimized using Equation~\ref{eq:grpo}.

\subsection{Population Evolution}
\label{app:population-details}

\paragraph{Population.}
SPO maintains role-specific populations as the memory of executable programs
discovered so far.  In each round, these populations provide parent context for
mutation and crossover, and are then updated by selecting survivors from the
union of existing programs and newly generated executable candidates.  Keeping
a bounded population, rather than only the single best program, preserves
alternative executable designs that can seed later generations.  We run 30
program-discovery rounds and keep at most 10 executable programs per role.

\paragraph{Candidate generation and survivor selection.}
In the first round, the destroy and repair populations are initialized without
parent programs.  In subsequent rounds, candidates are generated primarily
through mutation and crossover, using the strategies and source code of
programs in the current populations as context.  If insufficient eligible
parents are available, the corresponding generation slots fall back to
parent-free generation.

Programs that fail the executable-program gates are excluded before population
selection.  Destroy programs are deduplicated by whitespace-normalized source
code.  Repair programs are deduplicated by both normalized source code and the
destroy-strategy context used for their generation.  Survivor selection first
retains the candidate with the highest population-selection score and then
greedily selects subsequent candidates according to
\[
s(p) + 0.1 \min_{q \in S} d(p,q),
\]
where \(s(p)\) is the population-selection score, \(S\) is the set of already
selected survivors, and \(d(p,q)\) is the Jaccard distance between the token
sets of their strategy descriptions.

\paragraph{Repair cross-play selection.}
Repair population selection follows the cross-play evaluation described in the
main text.  Destroy survivors are selected first and used to construct a common
leader panel.  For TSP, this panel contains up to five evolved destroy
programs; for CVRP, it uses a hybrid panel with up to two evolved destroy
programs and five fixed classical destroy operators.  Each repair candidate is
evaluated against the same panel, and the resulting common-panel score
\(C_R^{\mathrm{pop}}\) is used for repair survivor selection.  This population
score is separate from the destroy-conditional repair credit used for GRPO
updates.
\section{Problem Settings and Operator Interfaces}
\label{app:problems}

This section specifies the problem instances, reference solutions, initial
solutions, operator interfaces, and validity requirements used in our
experiments.

\subsection{Traveling Salesperson Problem}
\label{app:tsp-setting}

\paragraph{Problem definition.}
For a Euclidean TSP instance with coordinates
\(\{c_i\}_{i=1}^{n}\subset\mathbb{R}^{2}\), a solution is a permutation
\(\sigma\) of the \(n\) nodes. Defining \(\sigma_{n+1}=\sigma_1\), its objective
value is

\begin{equation}
f(\sigma)
=
\sum_{i=1}^{n}
\left\lVert c_{\sigma_i}-c_{\sigma_{i+1}}\right\rVert_2.
\end{equation}

\paragraph{Data and reference solutions.}
Synthetic coordinates are sampled independently and uniformly from
\([0,1)^2\). Discovery uses six 50-node instances, with two stochastic LNS
rollouts per instance for candidate evaluation. The held-out synthetic test
groups reported in Table~\ref{tab:main_results} contain 32 instances per group.
Evaluation additionally uses TSPLIB instances, grouped by instance size as in
Table~\ref{tab:main_results}. Synthetic reference tour lengths are obtained
with Concorde~\citep{applegate2011tsp}, whereas TSPLIB uses the reported
optimal tour lengths~\citep{reinelt1991tsplib}. Benchmark distances and
reference values are computed using matching edge-weight conventions.

\paragraph{Initialization and operator interfaces.}
Each LNS rollout starts from a seeded random tour. The destroy program has the
interface

\begin{verbatim}
destroy(dist, current_tour, steps_since_improvement, rng)
    -> (partial_tour, removed_nodes)
\end{verbatim}

where \texttt{dist} is the distance matrix, \texttt{current\_tour} is the
current complete tour, and \texttt{rng} is the supplied random-number
generator. The returned destruction may remove at most \(35\%\) of the nodes.

The repair program has the interface

\begin{verbatim}
repair(dist, partial_tour, removed_nodes,
       steps_since_improvement, rng)
    -> complete_tour
\end{verbatim}

and must return a complete valid tour.

\subsection{Capacitated Vehicle Routing Problem}
\label{app:cvrp-setting}

\paragraph{Problem definition.}
A CVRP instance contains a depot \(0\), customers
\(\{1,\ldots,n\}\) with demands \(q_i\), and vehicle capacity \(Q\). A feasible
solution consists of depot-returning routes such that every customer appears in
exactly one route and the total demand of every route does not exceed \(Q\).
The objective is the sum of route lengths, including travel from and back to
the depot.

\paragraph{Data and reference solutions.}
For synthetic instances, the depot is located at \((0.5,0.5)\), customer
coordinates are sampled independently and uniformly from \([0,1)^2\), demands
are sampled uniformly from the integers \(\{1,\ldots,9\}\), and \(Q=50\).
Discovery uses six 50-customer instances. The synthetic test groups reported
in Table~\ref{tab:main_results} use the corresponding held-out instances at
each evaluation size.

Synthetic reference costs are produced by one PyVRP run per instance with a
fixed random seed and a size-dependent time limit: 10 seconds for \(n=50\),
20 seconds for \(n=100\), 30 seconds for \(n=150\), and 60 seconds for
\(n=200\)~\citep{wouda2024pyvrp}. These values are heuristic references rather
than certified optima.
Evaluation additionally
uses the A, B, P, and X families of CVRPLIB
\citep{augerat1995computational,uchoa2017benchmark}, with their corresponding
benchmark reference costs. We therefore report reference gaps rather than
optimality gaps for all groups.

\paragraph{Initialization and operator interfaces.}
Each LNS rollout starts from a seeded random capacity-feasible solution. Routes
contain customer indices only, with depot \(0\) implicit at both ends. The
destroy program has the interface

\begin{verbatim}
destroy(dist, demands, capacity, current_routes,
        steps_since_improvement, rng)
    -> (partial_routes, removed_customers)
\end{verbatim}

and may remove at most \(35\%\) of the customers. The repair program has the
interface

\begin{verbatim}
repair(dist, demands, capacity, partial_routes,
       removed_customers, steps_since_improvement, rng)
    -> complete_routes
\end{verbatim}

and must return a complete capacity-feasible solution.

\section{Baseline Implementations and Evaluation Budgets}
\label{app:evaluation-protocol}

We group baselines into handcrafted methods, test-time search methods, and
program-discovery methods. For each family, we specify both its test-time
procedure and, where applicable, its offline discovery budget.

\paragraph{Handcrafted baselines.}
Nearest Neighbor (NN) and Farthest Insertion (FI) for TSP, and
Clarke--Wright savings (CW) for CVRP, each perform a single construction
without restarts.

For TSP, 2-opt starts from the NN solution and terminates when no improving
move remains, with at most \(n^2\) accepted improving moves. The 3-opt baseline
starts from the NN+2-opt solution and performs at most 500 additional
improving moves. For CVRP, local search starts from the CW solution and applies
intra-route 2-opt, relocate, and swap moves for at most 500 improving moves.

\paragraph{Test-time search baselines.}
ALNS \citep{ropke2006alns} directly searches on each test instance using its
own adaptive operator-selection procedure. It is evaluated for 500
destroy--repair iterations per instance and evaluation seed, and we report the
best solution encountered during the rollout.

For CVRP, OR-Tools is run directly on each test instance with
size-dependent solver search limits selected to approximately match the
wall-clock time of a 500-iteration SPO LNS run at the corresponding scale:

\begin{center}
\small
\begin{tabular}{@{}c c @{\qquad} c c@{}}
\toprule
Instance size & Limit & Instance size & Limit \\
\midrule
\(n \leq 20\)        & \(1\) s   & \(20 < n \leq 50\)   & \(2\) s \\
\(50 < n \leq 100\)  & \(5\) s   & \(100 < n \leq 150\) & \(8\) s \\
\(150 < n \leq 200\) & \(10\) s  & \(200 < n \leq 500\) & \(30\) s \\
\(500 < n \leq 750\) & \(60\) s  & \(n > 750\)          & \(120\) s \\
\bottomrule
\end{tabular}
\end{center}

The limit applies to solver search rather than total end-to-end execution
time.

\paragraph{Program-discovery baselines.}
These methods first obtain destroy--repair programs offline and then evaluate
the selected programs for 500 destroy--repair iterations per instance and
evaluation seed, reporting the best solution encountered during each rollout.

For G-LNS \citep{zhao2026glns}, we directly use the operator programs and
implementation released by its authors; we do not retrain the method or assign
it a new discovery budget.

EoH \citep{liu2024eoh}, FunSearch \citep{romeraparedes2024funsearch}, and
ReEvo \citep{ye2024reevo} are general automatic program-discovery frameworks.
We implement them using the same problem-specific destroy--repair interfaces
and LNS runner as SPO.
Each method is assigned a nominal budget of 13,950 jointly generated
destroy--repair candidates, corresponding to 27,900 operator programs and
matching SPO's nominal total of 900 destroy and 27,000 repair generations.
Each candidate is evaluated using two 100-step LNS rollouts with different
random seeds. Runs are stopped once their discovery trajectories plateau, so
this budget is an upper bound rather than the realized candidate count.
Final candidates are selected through periodic 500-step validation evaluations,
with time limits of 10 seconds for TSP and 30 seconds for CVRP.
\section{Program Generation Contexts and Prompts}
\label{app:prompts}

This appendix describes the serialized generation contexts used by the
role-specific LLM generators introduced in Section~\ref{sec:optimization}.  A
generation context is not a single fixed prompt.  It is assembled from a
shared task contract and, depending on the discovery stage, additional
population-derived context.

The shared contract specifies the optimization problem, the destroy or repair
operator interface, feasibility requirements, and
the required output format: a concise strategy statement followed by one
executable Python function.  We refer to this shared contract as the basic
generation prompt.  During initialization, the same-role parent context is
empty, so the generator samples a new program directly from this basic
contract.  During evolutionary discovery, the same contract is retained, but
the context is augmented with parent strategies and source programs sampled
from the current population.

For destroy generation, the context corresponds to \(\gamma_D\).  For repair
generation, the context corresponds to \(\gamma_R\) and additionally includes
destroy-specific information for the paired destroy operator \(D_i\).  In the
initialization templates below, this destroy-conditioning information is
exposed through the placeholder \texttt{destroy\_strategy}.  In evolutionary
rounds, the serialized context records whether the paired destroy information
and the same-role parent information are supplied as strategy text, source
code, or both.

\subsection{Basic Initialization Prompts}
\label{app:initialization-prompts}

The following prompt boxes reproduce the basic initialization templates used
for TSP and CVRP destroy and repair generation.  These templates contain the
complete callable contract and requested program behavior for each operator
role.  Later mutation and crossover prompts reuse the same role-specific
contract, but append the evolutionary context described in
Section~\ref{app:evolution-prompts}.

\begin{promptbox}{TSP Destroy Initialization Prompt}
You are an expert in designing adaptive destroy operators for TSP Large Neighborhood Search. Discover a reusable adaptive destroy operator for Large Neighborhood Search on the Traveling Salesman Problem. The operator should create meaningful perturbations that enable effective reconstruction.

Inputs: distance_matrix (distance_matrix[i][j]: the distance from node i to node j), current_solution, steps_since_improvement, and rng.
Output: partial_solution and removed_nodes.

Design one adaptive removal policy that identifies meaningful removal regions from the current tour and uses steps_since_improvement to change concrete removal decisions rather than applying a fixed policy.

Destroy requirements:
- removed_nodes must contain node ids from current_solution
- construct removed_nodes before partial_solution
- partial_solution must preserve the relative order of current_solution
- every node must appear exactly once
- never remove more than 35% of the tour

Scale-aware guidance: derive removal size and candidate scope from the tour size, favor local structural evaluation over exhaustive search, and avoid fixed thresholds tied to a particular problem scale.

Output exactly two parts.
STRATEGY: one concise sentence naming the exact decision the code will implement, not a broader idea.
CODE: exactly one Python function. The function must implement the main mechanism named in STRATEGY.
\end{promptbox}

\begin{promptbox}{TSP Repair Initialization Prompt}
You are an expert in designing adaptive repair operators for TSP Large Neighborhood Search. Discover a reusable adaptive repair operator for Large Neighborhood Search on the Traveling Salesman Problem. The operator should reconstruct a complete tour from the destroy output.

Inputs: distance_matrix (distance_matrix[i][j]: the distance from node i to node j), partial_solution, removed_nodes, steps_since_improvement, and rng.
- partial_solution is shorter than the complete tour; use len(partial_solution) for current insertion positions and len(distance_matrix) only as the total node count.
Output: repaired_tour (the complete tour after reinserting all removed nodes).

Use this paired destroy strategy as context when designing the repair:
{destroy_strategy}

Design one reconstruction policy that complements the destroy result by using the partial solution and removed nodes as reconstruction context. steps_since_improvement may adjust insertion order, candidate scope, or tie handling when useful.

Repair requirements:
- reinsert every removed node exactly once
- preserve the relative order of retained nodes
- use the retained tour only as insertion context
- do not perform additional tour optimization

Scale-aware guidance: derive insertion candidate scope from the current tour size and the number of removed nodes, favor local insertion evaluation over exhaustive search, and avoid fixed thresholds tied to a particular problem scale.

Output exactly two parts.
STRATEGY: one concise sentence naming the exact decision the code will implement, not a broader idea.
CODE: exactly one Python function. The function must implement the main mechanism named in STRATEGY.
\end{promptbox}

\begin{promptbox}{CVRP Destroy Initialization Prompt}
You are an expert in designing adaptive destroy operators for CVRP Large Neighborhood Search. Discover a reusable adaptive destroy operator for Large Neighborhood Search on the Capacitated Vehicle Routing Problem. The operator should remove customers while preserving route order and capacity feasibility of the retained partial routes.

Inputs: distance_matrix, demands, capacity, current_routes, steps_since_improvement, and rng.
- Node 0 is the depot and never appears inside current_routes.
- demands[0] is zero. Every customer node has positive demand.
Output: partial_routes and removed_customers.

Design one adaptive removal policy that selects meaningful customers or route regions and uses steps_since_improvement to change concrete removal decisions rather than applying a fixed policy.

Destroy requirements:
- removed_customers must contain customer ids from current_routes
- construct removed_customers before partial_routes
- partial_routes must preserve the relative order of retained customers within each route
- every customer must appear exactly once across partial_routes plus removed_customers
- never remove depot 0
- never remove more than 35% of customers
- retained partial routes must remain capacity-feasible

Scale-aware guidance: derive removal size and candidate route scope from the route count, customer count, and capacity pressure; favor local route/customer evaluation over exhaustive search; avoid fixed thresholds tied to a particular CVRP size or vehicle count.

Output exactly two parts.
STRATEGY: one concise sentence naming the exact decision the code will implement, not a broader idea.
CODE: exactly one Python function. The function must implement the main mechanism named in STRATEGY.
\end{promptbox}

\begin{promptbox}{CVRP Repair Initialization Prompt}
You are an expert in designing adaptive repair operators for CVRP Large Neighborhood Search. Discover a reusable adaptive repair operator for Large Neighborhood Search on the Capacitated Vehicle Routing Problem. The operator should reconstruct complete feasible routes from the destroy output.

Inputs: distance_matrix, demands, capacity, partial_routes, removed_customers, steps_since_improvement, and rng.
- Node 0 is the depot and never appears inside routes.
- Each returned route must have total demand <= capacity.
Output: repaired_routes.

Use this paired destroy strategy as context when designing the repair:
{destroy_strategy}

Design one reconstruction policy that complements the destroy result by reinserting all removed customers into feasible routes. steps_since_improvement may adjust insertion order, candidate scope, or tie handling when useful.

Repair requirements:
- reinsert every removed customer exactly once
- every original retained customer must remain exactly once
- never include depot 0 inside returned routes
- every returned route load must be <= capacity
- preserve retained route order whenever feasible
- do not perform additional global route optimization after insertion

Scale-aware guidance: derive feasible insertion candidate scope from the current route count, customer count, removed-customer count, and residual capacities; favor bounded capacity-feasible insertion evaluation over exhaustive search; avoid fixed thresholds tied to a particular CVRP size or vehicle count.

Output exactly two parts.
STRATEGY: one concise sentence naming the exact decision the code will implement, not a broader idea.
CODE: exactly one Python function. The function must implement the main mechanism named in STRATEGY.
\end{promptbox}

\subsection{Evolutionary Prompt Forms}
\label{app:evolution-prompts}

During program discovery, SPO maintains separate destroy and repair
populations.  The first round initializes these populations using prompts
without same-role parents.  In later rounds, new candidates are generated from
the current population through mutation or crossover.  These evolutionary
prompts are constructed by augmenting the basic destroy or repair prompt with
population-derived parent context, while keeping the same operator interface,
feasibility requirements, and output format.

For mutation, the prompt includes one parent strategy and source program,
together with an instruction describing the intended type of modification.
For crossover, the prompt includes two parent strategies and source programs,
and asks the generator to produce a new program using the parents as design
context.  The resulting candidates are then evaluated and used by the training
and population-update procedure described in Appendix~\ref{app:implementation}.

Table~\ref{tab:evolution-strategies} summarizes the evolutionary prompt forms.
The table describes the instruction type rather than reproducing the full
serialized prompt text.

\begin{table}[h]
\caption{Evolutionary prompt forms used during program discovery.}
\label{tab:evolution-strategies}
\centering
\footnotesize
\setlength{\tabcolsep}{4pt}
\renewcommand{\arraystretch}{1.08}
\begin{tabular}{@{}p{1.35in}p{0.65in}p{2.95in}@{}}
\toprule
Prompt form & Parents & Instruction \\
\midrule
Mechanism replacement & One & Replace one major decision mechanism while
preserving the remainder of the parent design. \\
State-dependent control redesign & One & Preserve the main mechanism but alter
thresholds, budgets, candidate scopes, or adaptation schedules. \\
Simplification & One & Remove redundant or unnecessarily complex components
without introducing a new major mechanism. \\
Divergent crossover & Two & Use the parents as records of explored ideas and
generate a substantially different heuristic. \\
Shared-principle crossover & Two & Extract a shared high-level principle and
implement it through a new program structure. \\
\bottomrule
\end{tabular}
\end{table}

\section{LNS-State Intervention Details}
\label{app:state-intervention}

Figure~\ref{fig:optimization-dynamics}(b1) evaluates a fixed discovered TSP
operator pair while modifying only the LNS-state signal supplied during
execution. The interventions are defined as follows:

\begin{itemize}
    \item \textbf{Fixed \(k\):} set \(z_t=k\) throughout the rollout, with
    \(k\in\{0,5,10\}\);
    \item \textbf{Shuffled:} randomly permute the state sequence observed in
    the corresponding unmodified rollout;
    \item \textbf{Inverted:} set \(z_t=30\) when the observed state is below
    \(10\), and set \(z_t=0\) otherwise.
\end{itemize}

For each instance--seed pair \(u\) and intervention \(a\), we compute the
paired change in the final best reference gap,

\begin{equation}
\Delta^{u}(a)
=
g^{u}_{(a)}-g^{u}_{(\mathrm{base})},
\end{equation}

where \(g^{u}_{(\mathrm{base})}\) denotes the result obtained using the
unmodified state signal. Figure~\ref{fig:optimization-dynamics}(b1) reports
the mean paired difference and its \(95\%\) confidence interval, computed from
2,000 bootstrap resamples of the instance--seed pairs.

\end{document}